%% file: root.tex
\documentclass[letterpaper, 10 pt, conference]{ieeeconf}  

\IEEEoverridecommandlockouts                              

\usepackage{amsfonts}
\usepackage{graphicx}
\usepackage{amsmath}
\usepackage{cleveref}
\usepackage{makecell}
\usepackage{algorithm}
\usepackage{algorithmic}
\usepackage{adjustbox}
\usepackage{multirow}
\usepackage{subcaption}
\usepackage{tabularx}
\usepackage[table]{xcolor}
\usepackage{booktabs}       
\usepackage{wrapfig}
\title{\LARGE \bf
Info3R: Information-Adaptive Test-Time Training for 3D Reconstruction
}

\author{Sunghyun Baek$^{1}$, Hanna Bae$^{1}$, Minchan Kwon$^{1}$, and Junmo Kim$^{1}$
\thanks{$^{1}$The authors are with the Department of Electrical Engineering, Korea Advanced Institute of Science and Technology, South Korea. email: \{baeksh, hannabae, kmc0207, junmo.kim\}@kaist.ac.kr}%
}

\begin{document}

\maketitle
\thispagestyle{empty}
\pagestyle{empty}

\begin{abstract}
Transformer-based models have recently achieved strong performance on 3D reconstruction from images, and recent works extend them to process video streams in an online manner for real-world deployment. However, existing methods overlook two key signals when handling long image streams: the importance of each incoming frame and the information saturation of the model's internal state. In this paper, we propose Info3R, a novel information-adaptive test-time training method for the online 3D reconstruction. We introduce an information-aware state update that modulates the state update strength based on the redundancy and informativeness of each incoming frame. To restore the state's plasticity—its capacity to incorporate new observations—we propose a dynamic state reset, triggered by the cumulative magnitude of state updates and the model's prediction confidence and accompanied by an anchor-to-world alignment. Our method achieves consistent improvements on camera pose estimation, video depth estimation, and 3D reconstruction, while substantially mitigating the performance degradation in the long sequence evaluation. Notably, on KITTI Odometry, our method achieves on average 1.68× lower ATE than LongStream, demonstrating its robustness on extended outdoor sequences.
\end{abstract}

\section{INTRODUCTION}
\label{sec:3d intro}
Reconstructing 3D geometry from images is a core technology for applications such as robotics, AR/VR, and autonomous driving. Recently, transformer-based feed-forward models (e.g., VGGT \cite{vggt}, $\pi^3$ \cite{pi3}) that jointly process multiple images have attracted significant attention for their outstanding performance. Beyond reconstructing isolated scenes, the ability to process long image streams in an online manner has emerged as a key requirement for real-world deployment. 
Driven by this demand, streaming-based 3D reconstruction methods \cite{ttt3r, longstream} have been proposed.

Several works \cite{ttt3r, stream3r, streamingvggt, longstream, long3r} have been explored to extend 3D reconstruction to long sequences. 
Specifically, TTT3R \cite{ttt3r} reinterprets the state update of CUT3R \cite{cut3r}, an online 3D perception model, through the lens of test-time training and improves it in a training-free manner.
However, TTT3R has the following limitations. First, the importance of each frame is not taken into account when updating the model state, causing redundant or degraded frames to be integrated with the same intensity as informative ones and ultimately degrading reconstruction quality over long horizons. Second, the accumulation of scene information in an online manner leads to information saturation, which reduces the model's plasticity to incorporate new observations. This issue, combined with the fixed first-frame anchor problem in long sequence settings, calls for a principled reset mechanism that balances plasticity and global consistency.
\begin{figure}[t]
  \centering
\includegraphics[width=\linewidth]{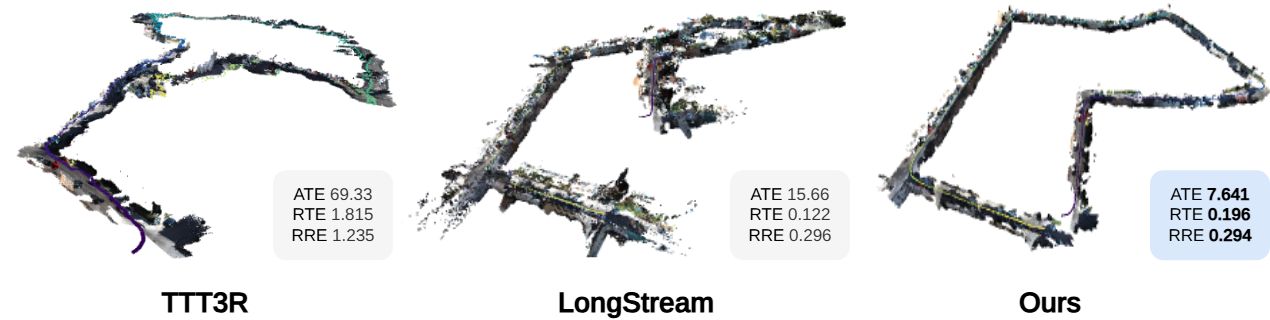}
  \caption{
  \textbf{3D reconstruction on KITTI Odometry Seq.~07.} TTT3R fails to estimate the overall trajectory, and LongStream improves over it but fails at loop closure when revisiting the same location. In contrast, Info3R reconstructs a coherent and geometrically consistent scene, demonstrating robustness to long-horizon streaming inputs.}
  \label{fig:visualize}
  \vspace{-4mm}
\end{figure}
To address these limitations, we propose Info3R, an information-adaptive test-time training method that improves the state update process of CUT3R.
First, we introduce \textit{Information-Aware State Update} that modulates the adaptive learning rate of state tokens. 
Specifically, we quantify both the information redundancy and the informativeness of the current frame, and combine them into an importance score that modulates the learning rate of the state update. 
Next, we propose \textit{Dynamic State Reset} that addresses the information saturation inherent to finite-capacity state.
We detect the information saturation of state by tracking the cumulative learning rate alongside the prediction confidence, and reset the state whenever this combined signal exceeds a threshold.
Subsequent predictions are then produced in the anchor's coordinate system and mapped back to the world coordinate via an anchor-to-global coordinate alignment. Jointly using these mechanisms enables stable and accurate 3D perception across long sequences.

We validate Info3R on a diverse set of 3D perception tasks, including camera pose estimation, video depth estimation, and 3D reconstruction.
Our training-free approach achieves a $1.68\times$ lower ATE than LongStream (30.48 vs.\ 51.24) on KITTI Odometry and consistently outperforms prior methods across the other benchmarks.
Notably, our advantage over the baselines grows with sequence length, indicating that our method effectively mitigates the performance degradation observed in extended streams.
We further provide an observation that the world coordinate confidence decreases with information saturation, manifested as an increase in the rank of state tokens, supporting the necessity of our method.

\section{RELATED WORK}
\label{sec:rel works}
\subsection{Feed-forward 3D Reconstruction Methods}
Recent advancements in 3D reconstruction have shifted from traditional pipelines \cite{bundle, sfm, t_visualmodeling}—which separately estimate camera poses, intrinsics, and correspondences—toward end-to-end feed-forward architectures. 
DUSt3R \cite{wang2024dust3r} pioneered this shift by directly predicting 3D point maps from image pairs, bypassing the complexities of conventional Structure-from-Motion (SfM) and Multi-View Stereo (MVS). Building upon this, MASt3R \cite{mast3r} integrated a dense local feature head and a matching loss to enhance 3D-aware image matching performance. 
VGGT \cite{vggt} introduced a visual geometry foundation model that jointly predicts camera parameters, depth, point maps, and tracks across multiple views in a single forward pass. 
Expanding on this, $\pi^3$\cite{pi3} proposed a permutation-equivariant model that eliminates the dependency on a fixed reference view, providing a more flexible approach.

\subsection{Streaming 3D Reconstruction Methods}
As the demand for real-time applications grows, sequential processing of incoming image streams has become a critical research area. CUT3R \cite{cut3r} serves as a representative baseline for recurrent 3D reconstruction, maintaining a persistent state token to process streaming data online. To address the performance degradation observed in extended evaluations, TTT3R \cite{ttt3r} reformulated the state update as a test-time training (TTT) process. 
Inspired by the success of Large Language Models (LLMs), several recent works have adapted transformer architectures for efficient geometry processing. 
Stream3R \cite{stream3r} utilizes a causal transformer with cached memory tokens to handle streaming inputs. Similarly, StreamVGGT \cite{streamingvggt} modifies the attention mechanism of VGGT into a causal structure, leveraging KV caching to reduce memory overhead. Finally, to ensure stability over ultra-long sequences, LongStream \cite{longstream} removed the dependence on the first-frame anchor and introduced keyframe-relative poses, orthogonal scale learning, and periodic refreshes, enabling consistent geometry estimation in extended outdoor environments.


\section{BACKGROUND}
\label{sec:backgrounds}

\subsection{Multi-view 3D Reconstruction}
Given an image stream $\{\mathbf{I}_t\}_{t=0}^{T}$ captured from different viewpoints, where $t$ denotes the time step, the objective of multi-view 3D reconstruction is to estimate the 3D geometry and camera poses for the entire sequence.
For each frame $\mathbf{I}_t$ within the stream, the model predicts 3D point maps $\hat{\mathbf{X}}_t \in \mathbb{R}^{H \times W \times 3}$ and confidence maps $\mathbf{C}_t \in \mathbb{R}^{H \times W}$ in two distinct coordinate systems: the local camera coordinate and the global world coordinate.

\subsection{Continuous 3D Perception Model}
Continuous 3D Perception Model (CUT3R)~\cite{cut3r} processes an image stream sequentially in an online manner using a set of state tokens $\mathbf{S}_t \in \mathbb{R}^{M \times D}$.
At each time step $t$, the current image $\mathbf{I}_t$ is encoded into image tokens $\mathbf{F}_t \in \mathbb{R}^{N \times D}$ via a Vision Transformer (ViT) encoder~\cite{vit}.
Here, $M$, $N$, $D$ denote the number of state and image tokens, and the embedding dimension, respectively.
Through interconnected transformer decoders, the image tokens and the previous state $\mathbf{S}_{t-1}$ exchange information via cross-attention, yielding concurrent state-readout and state-update:
\begin{equation}
\label{eq:interaction}
    [\mathbf{z}'_t, \mathbf{F}'_t], \mathbf{S}_t = \text{Decoders}([\mathbf{z}, \mathbf{F}_t], \mathbf{S}_{t-1}),
\end{equation}
where $\mathbf{z}'_t$, $\mathbf{F}'_t$, and $\mathbf{S}_t$ denote the updated pose token, image tokens, and state tokens, respectively.
From these updated features, three task-specific heads extract 3D representations and camera poses:
\begin{align}
\hat{\mathbf{X}}^{\text{self}}_t, \mathbf{C}^{\text{self}}_t
    &= \text{Head}_{\text{self}}(\mathbf{F}'_t), \label{eq:head_self}\\
\hat{\mathbf{X}}^{\text{world}}_t, \mathbf{C}^{\text{world}}_t
    &= \text{Head}_{\text{world}}(\mathbf{F}'_t, \mathbf{z}'_t), \label{eq:head_world}\\
\hat{\mathbf{T}}_{A_0 \leftarrow V_t}
    &= \text{Head}_{\text{pose}}(\mathbf{z}'_t). \label{eq:head_pose}
\end{align}
where $\hat{\mathbf{X}}^{\text{self}}_t$ and $\hat{\mathbf{X}}^{\text{world}}_t$ denote reconstructed 3D point maps, each associated with confidence maps $\mathbf{C}_t^\text{self}$ and $\mathbf{C}_t^\text{world}$. 
Specifically, these represent point maps in the current view coordinate $V_t$ and the world coordinate $A_0$ (defined as the first frame).
The predicted pose $\hat{\mathbf{T}}_{A_0 \leftarrow V_t} \in SE(3)$ represents the camera's relative pose. It also transforms coordinates from the current view $V_t$ to the world coordinate $A_0$.

\subsection{Test-Time Training for 3D Reconstruction}
Although pre-trained models such as CUT3R generalize well, their performance tends to degrade on long image streams. 
To mitigate this, Test-Time Training for 3D Reconstruction (TTT3R) \cite{ttt3r} reinterprets the state-update mechanism as a test-time training process over the state tokens. 
A per-token learning rate is then derived from the spatial alignment confidence
\(
\beta_t = \sigma\!\left(\sum_m \mathbf{Q}_{\mathbf{S}_{t-1}} \mathbf{K}_{\mathbf{F}_t}^{\top}\right),
\)
where $\mathbf{Q}_{\mathbf{S}_{t-1}}$ is a computed query from the current state tokens, $\mathbf{K}_{\mathbf{F}_t}$ and $\mathbf{V}_{\mathbf{F}_t}$ from the current image tokens, $m$ indexes the attention heads, and $\sigma(\cdot)$ is the sigmoid activation.
Denoting the decoder's output state in \Cref{eq:interaction} as $\hat{\mathbf{S}}_t$, the state is updated by interpolating between the previous state $\mathbf{S}_{t-1}$ and the new candidate $\hat{\mathbf{S}}_t$ 
via the per-token learning rate:
\begin{equation}
\mathbf{S}_t = \mathbf{S}_{t-1} + \beta_t \odot (\hat{\mathbf{S}}_t - \mathbf{S}_{t-1}).
\label{eq:ttt3r_update}
\end{equation}
This can be viewed as a generalization of CUT3R, where $\beta_t$ is implicitly fixed to $\mathbf{1}$.

\section{METHOD}
\label{sec:method}
\begin{figure}[t]
  \centering
  \resizebox{\linewidth}{!}{
  \includegraphics[]{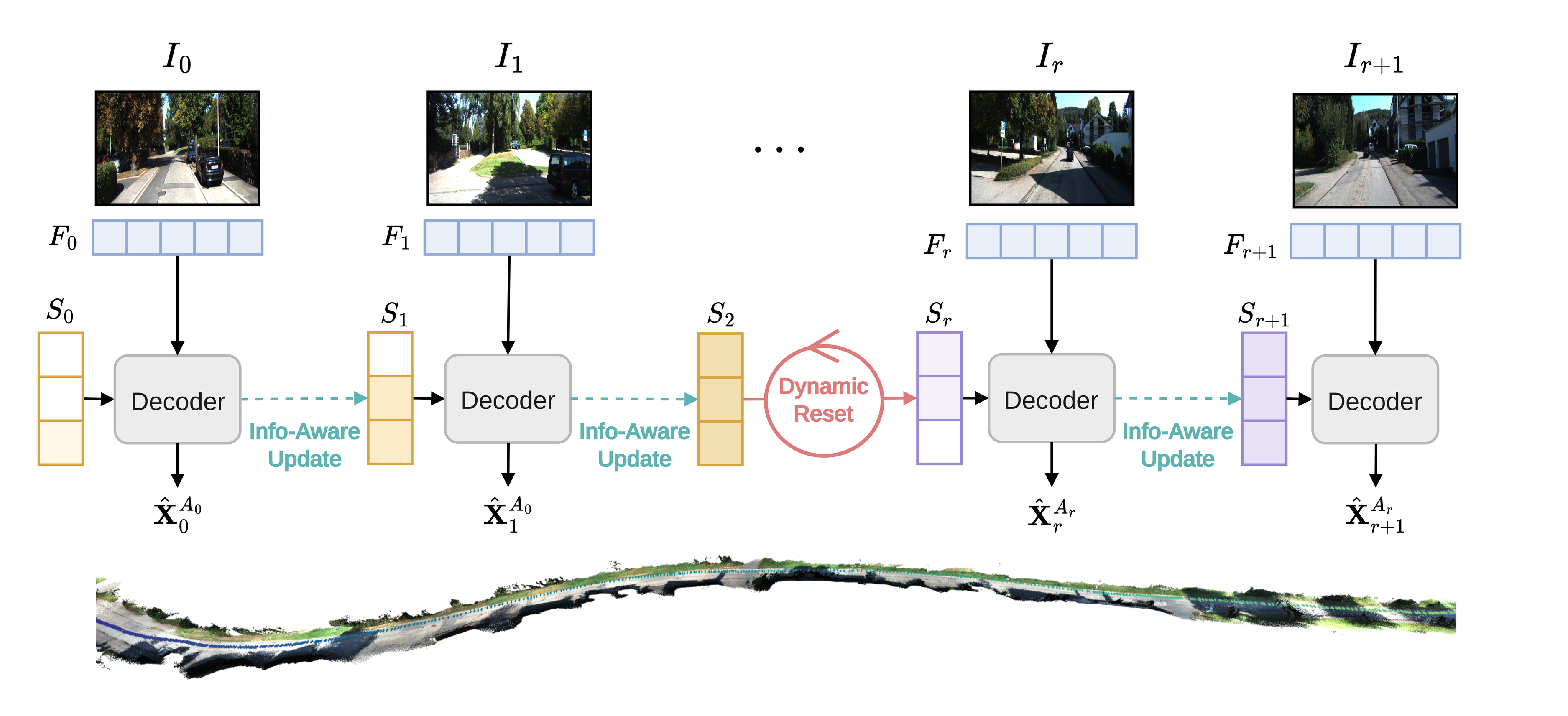}
  }
  \caption[Overview of Info3R]{\textbf{Overview of Info3R.} 
For each incoming frame, the Information-Aware State Update modulates the state update with an adaptive learning rate $\alpha_t$ derived from information redundancy and informativeness.
The Dynamic State Reset re-initializes the saturated state when the trigger metric $\Gamma_t$ exceeds a threshold, and preserves global geometry through anchor-to-world alignment.
  }
  \label{fig:overview}
  \vspace{-5mm}
\end{figure}

We propose Info3R, an information-adaptive test-time training framework 
for 3D reconstruction. It dynamically updates state tokens based on each 
frame's significance, accumulating only meaningful information from long 
image streams.
As illustrated in \Cref{fig:overview}, Info3R consists of two complementary mechanisms.
Information-Aware State Update (\Cref{sec:information aware state update}) modulates the learning rate based on the redundancy and informativeness of each frame, so that novel and reliable observations are integrated while redundant or degraded ones are suppressed. Dynamic State Reset (\Cref{sec:dynamic state reset}) refreshes the saturated state when saturation is detected and preserves global geometry through coordinate alignment, restoring plasticity without sacrificing global structure. The details of each component are described in the following sections.

\subsection{Information-Aware State Update}
\label{sec:information aware state update}
To selectively accumulate meaningful information to the state, we modulate the learning rate using two complementary criteria: information redundancy and informativeness. The former suppresses updates from largely overlapping frames, leaving room in the finite state capacity for novel observations; the latter scales the update by the structural richness of the current frame, preventing uninformative or noisy views from contaminating the state. Combining both metrics yields an adaptive learning rate that accumulates novel and reliable information.

\noindent\textbf{Information redundancy term.}
To quantify the overlap between consecutive frames, we define a redundancy-aware weight, $w_{\text{redundant}}$, based on the similarity of spatial token representations.
Given consecutive image tokens $\mathbf{F}_t, \mathbf{F}_{t-1}$, let $\mathbf{f}_t^{(i)} \in \mathbb{R}^D$ denote the $i$-th spatial token of $\mathbf{F}_t$.
The redundancy-aware weight is defined as follows:
\begin{equation}
\label{eq:w_redundant}
w_{\text{redundant}} = 1 - \frac{1}{N} \sum_{i=1}^{N} \frac{\mathbf{f}_t^{(i)} \cdot \mathbf{f}_{t-1}^{(i)}}{\|\mathbf{f}_t^{(i)}\|_2 \,\|\mathbf{f}_{t-1}^{(i)}\|_2}.
\end{equation}

\noindent\textbf{Informativeness term.} 
While the redundancy-aware weight encourages the accumulation of novel information in the state tokens, it has limitations in practical scenarios.
For instance, when a severe motion blur occurs during video capture, the feature differences between consecutive image tokens may significantly increase despite the absence of meaningful geometric information.
This leads to an inappropriately high update weight, potentially contaminating the state tokens. 
To alleviate this issue, we assess the structural richness of each frame beyond information redundancy. 
Specifically, we apply singular value decomposition (SVD) to the image tokens $\mathbf{F}_t \in \mathbb{R}^{N \times D}$, yielding singular values $\sigma_1 \geq \sigma_2 \geq \cdots \geq \sigma_n \geq 0$, where each $\sigma_k^2$ measures the energy along the $k$-th principal direction of the token features. 
The effective rank is then defined as the minimum number of components required to capture $95\%$ of the total energy:
\begin{equation}
r_\text{eff} = \min_{r \in \mathbb{N}} \left\{ r : \sum_{k=1}^{r} \sigma_k^2 \geq 0.95 \sum_{j} \sigma_j^2 \right\}.
\end{equation}
Intuitively, $r_\text{eff}$ reflects how many independent feature directions are needed to describe the frame: A higher effective rank implies greater feature diversity, which in turn correlates with richer, more informative content; a lower value indicates that features collapse onto few dominant directions, characteristic of texture-less or degraded observations.
The informativeness weight is defined as $w_{\text{info}} = \min(r_{\text{eff}} / \tau,\, 1)$, where $\tau$ is a normalization factor.

\noindent\textbf{State Update.}
The final learning rate scales a base learning rate $\alpha_{\text{base}}$, a hyperparameter controlling the overall update magnitude, by both weights:
\begin{equation}
\label{eq:adaptive learning rate}
\alpha_t = \min \left( \alpha_{\text{base}} \cdot w_{\text{redundant}} \cdot w_{\text{info}}, 1.0 \right).
\end{equation}
Using this adaptive learning rate, the state tokens $\mathbf{S}_t$ are updated:
\begin{equation}
\label{eq:update rule}
\mathbf{S}_t = \mathbf{S}_{t-1} + \alpha_t \cdot (\hat{\mathbf{S}}_t - \mathbf{S}_{t-1}).
\end{equation}
This state update selectively integrates informative observations into the state while preventing contamination from redundant or unreliable frames.

\subsection{Dynamic State Reset}
\label{sec:dynamic state reset}
Despite the information-aware state update, repeated state updates eventually degrade performance due to the finite capacity of state tokens.
To resolve this, we propose Dynamic State Reset, which adaptively resets the state while preserving global geometry through coordinate alignment.
We first describe the reset and alignment procedure, and then introduce the information-driven trigger that decides when to reset.

\noindent\textbf{State Reset and Coordinate Alignment.}
When the reset is triggered, we re-initialize the saturated state to restore its plasticity for new observations.
Concretely, at the $c$-th reset cycle, we discard the current state and re-initialize it from the current image tokens:
\begin{equation}
    \mathbf{S}_{t_c} \leftarrow \text{Init}(\mathbf{F}_{t_c}),
\end{equation}
where $t_c$ is the time step at which the $c$-th reset occurs and $\text{Init}(\cdot)$ denotes the same initialization module used at the start of the sequence. 
For clarity, we omit the initialization of the pose token, which follows the same procedure.
After the reset, all subsequent predictions from $\text{Head}_{\text{self}}$ and $\text{Head}_{\text{world}}$ are no longer expressed in the original world coordinate $A_0$, but in a new \textit{anchor coordinate} $A_c$ defined by the re-initialized state.

To retain the long-term geometry across resets, we accumulate it into a pose transformation that chains consecutive anchor coordinates:
\begin{equation}
    \label{eq:anchor chain}
    \hat{\mathbf{T}}_{A_0 \leftarrow A_c} = \hat{\mathbf{T}}_{A_0 \leftarrow A_{c-1}} \cdot \hat{\mathbf{T}}_{A_{c-1} \leftarrow A_c},
\end{equation}
where $\hat{\mathbf{T}}_{A_{c-1} \leftarrow A_c}$ is obtained from the model's pose prediction at the reset step, expressed in the previous anchor coordinate $A_{c-1}$.
Since $A_0$ is defined as the world coordinate, this transformation maps the current anchor back to the global frame.
Using this accumulated transformation, the anchor-coordinate predictions are mapped back to the canonical world coordinate as
\begin{equation}
\label{eq:l2w_alignment}
    \hat{\mathbf{X}}_t^{\text{world}} = \hat{\mathbf{T}}_{A_0 \leftarrow A_c} \circ \hat{\mathbf{X}}_t^{A_c}, \quad 
    \hat{\mathbf{T}}_{A_0 \leftarrow V_t} = \hat{\mathbf{T}}_{A_0 \leftarrow A_c} \cdot \hat{\mathbf{T}}_{A_c \leftarrow V_t},
\end{equation}
where $\hat{\mathbf{X}}_t^{A_c}$ denotes the world-head output expressed in the current anchor coordinate $A_c$, $\circ$ denotes the action of an $SE(3)$ transformation on 3D points, and $\cdot$ denotes the composition of two $SE(3)$ transformations.

\noindent\textbf{Information-Driven Trigger.}
As image streams exhibit varying dynamics, fixed-interval resets are suboptimal, necessitating an adaptive criterion. To this end, we jointly track the cumulative update magnitude $\sum \alpha_i$ since the last reset, which reflects the total influx of information into the state, and the average world coordinate confidence $\bar{C}^{\text{world}}_t$, which tends to collapse over long sequences.
We combine these two signals into a single trigger metric:
\begin{equation}
\label{eq:trigger}
\Gamma_t = \sum_{i=t_c}^{t} \frac{\alpha_i}{\bar{C}^{\text{world}}_i},
\end{equation}
where $t_c$ denotes the time step of the most recent reset.
When $\Gamma_t$ exceeds a threshold $\gamma$, a new cycle starts.
$\Gamma_t$ grows monotonically with both signals, so a reset is initiated whenever the state has saturated or its predictions have grown unreliable.
This input-adaptive criterion, combined with coordinate alignment, refreshes the state on demand while preserving the global geometry accumulated across cycles.
We provide the full pseudocode of our method in \Cref{alg:dynamic_reset}.

\renewcommand{\algorithmiccomment}[1]{\hfill $\slash\slash$ #1}
\begin{algorithm}[H]
\caption[Dynamic State Reset with Information-Adaptive Update]{Dynamic State Reset with Information-Adaptive Update}
\label{alg:dynamic_reset}
\begin{algorithmic}[1]
\STATE \textbf{Require:} Image stream $\{\mathbf{I}_t\}_{t=0}^{N}$; Network $f_\theta$; Reset threshold $\gamma$; Base learning rate $\alpha_{\text{base}}$.
\STATE \textbf{Ensure:} Global point maps $\{\hat{\mathbf{X}}_t^{\text{world}}\}$; Global poses $\{\hat{\mathbf{T}}_{A_0 \leftarrow V_t}\}$.
\STATE Extract initial feature $\mathbf{F}_0 = \mathrm{Encoder}(\mathbf{I}_0)$
\STATE $\mathbf{S}_0 \gets \mathbf{F}_0$, $\quad c \gets 0$, $\quad \Gamma \gets 0$
\STATE $\hat{\mathbf{T}}_{A_0 \leftarrow A_c} \gets \mathbf{I}_{4 \times 4}$
\FOR{$t = 1, \dots, N$}
    \STATE $\mathbf{F}_t = \mathrm{Encoder}(\mathbf{I}_t)$
    \STATE Compute $w_{\text{redundant}}$ and $w_{\text{info}}$ from $\mathbf{F}_t$ and $\mathbf{F}_{t-1}$
    \STATE $\alpha_t \gets \min(\alpha_{\text{base}} \cdot w_{\text{redundant}} \cdot w_{\text{info}},\, 1.0)$ 
    \IF{$\Gamma > \gamma$}

        \STATE Predict $\hat{\mathbf{T}}_{A_c \leftarrow V_t}$ via $f_\theta(\mathbf{F}_t, \mathbf{S}_{t-1})$
        \STATE $c \gets c + 1$
        \STATE $\hat{\mathbf{T}}_{A_0 \leftarrow A_c} \gets \hat{\mathbf{T}}_{A_0 \leftarrow A_{c-1}} \cdot \hat{\mathbf{T}}_{A_{c-1} \leftarrow V_t}$
        \STATE $\mathbf{S}_{t-1} \gets \mathbf{F}_t$, $\quad \Gamma \gets 0$
    \ENDIF
    \STATE Obtain $\hat{\mathbf{S}}_t, \hat{\mathbf{T}}_{A_c \leftarrow V_t}, \hat{\mathbf{X}}_t^{A_c}, \bar{C}_t^{\text{world}}$ via $f_\theta(\mathbf{F}_t, \mathbf{S}_{t-1})$
    \STATE $\mathbf{S}_t \gets \mathbf{S}_{t-1} + \alpha_t (\hat{\mathbf{S}}_t - \mathbf{S}_{t-1})$ \COMMENT{State update (Eq.~\ref{eq:update rule})}
    \STATE $\hat{\mathbf{X}}_t^{\text{world}} \gets \hat{\mathbf{T}}_{A_0 \leftarrow A_c} \circ \hat{\mathbf{X}}_t^{A_c}$
    \STATE $\hat{\mathbf{T}}_{A_0 \leftarrow V_t} \gets \hat{\mathbf{T}}_{A_0 \leftarrow A_c} \cdot \hat{\mathbf{T}}_{A_c \leftarrow V_t}$
    \STATE $\Gamma \gets \Gamma + \dfrac{\alpha_t}{\bar{C}_t^{\text{world}}}$ \COMMENT{Update trigger metric (Eq.~\ref{eq:trigger})}
\ENDFOR
\RETURN $\{\hat{\mathbf{X}}_t^{\text{world}}\}_{t=0}^{N},\, \{\hat{\mathbf{T}}_{A_0 \leftarrow V_t}\}_{t=1}^{N}$
\end{algorithmic}
\end{algorithm}

\section{EXPERIMENTS}
\subsection{Experimental Setup}

\noindent\textbf{Datasets.}
We evaluate Info3R across three distinct tasks: camera pose estimation, video depth estimation, and 3D reconstruction.
Following TTT3R \cite{ttt3r}, we generate long sequence datasets from these benchmarks for extended sequence evaluation.
For camera pose estimation, we use ScanNet \cite{scannet} and TUM dynamics \cite{tumdynamics} to assess ego-motion tracking in complex indoor environments. 
For video depth estimation, we use KITTI \cite{kitti} and Bonn \cite{bonn} to encompass a diverse range of scenarios, including both static and dynamic scenes in indoor and outdoor settings.
For 3D reconstruction, we employ 7-Scenes \cite{7scenes} and NRGBD \cite{nrgbd} to validate global geometry aggregation. 
Additionally, to evaluate the proposed method on long outdoor sequences, we conduct experiments on the KITTI Odometry \cite{kitti}, following LongStream \cite{longstream}.

\noindent\textbf{Evaluation Metrics.} 
For camera pose estimation, Absolute Trajectory Error ($\text{ATE}$) measures global consistency and long-term drift after $Sim(3)$ alignment.
Regarding video depth estimation, $\text{Abs Rel}$ provides a scale-invariant measure of relative error and $\delta < 1.25$ (Threshold Accuracy) indicates the proportion of reliable pixel predictions across diverse spatial contexts. 
For 3D reconstruction, Acc (Accuracy) and Comp (Completeness) measure geometric precision and scene coverage via point-to-surface distance, while NC (Normal Consistency) captures local geometric details via surface-normal alignment.

\noindent\textbf{Implementation Details.} 
Our model employs a ViT-Large backbone \cite{vit} with pretrained weights from the official CUT3R \cite{cut3r} implementation. All experiments are conducted on a single NVIDIA GeForce RTX 4090 GPU. 
We use a batch size of 1 and the hyperparameters of our method are provided in the Appendix \ref{sec:hyperparameters}.
 
\noindent\textbf{Baselines.} 
Following the extended sequence evaluations of TTT3R \cite{ttt3r}, we adopt CUT3R \cite{cut3r} and TTT3R as our primary baselines, and additionally compare against LongStream \cite{longstream} on the KITTI Odometry dataset. 

\subsection{Main Results}
\label{sec:3d main results}
\input{tables/0_long_camera_pose}
\input{tables/0_visualize_long_camera_pose}
\input{tables/1_camera_pose_tables}
\input{tables/2_video_depth_tables}
\noindent\textbf{Long sequence camera pose estimation.} 
In \Cref{tab:kitti_result}, our method achieves the best average ATE on KITTI Odometry compared to CUT3R, TTT3R, and LongStream. 
Although LongStream performs better on short sequences with simple ego-motion (Seq.~01, Seq.~03, and Seq.~04), our approach shows a significant performance margin in all other cases, including sequences with frequent rotational motion.
For example, on Seq.~00, our method recovers a trajectory that returns to its starting point without any explicit loop closure constraints, as shown in \Cref{fig:kitti traj error}. These results indicate that our method achieves strong long-sequence pose estimation even without the dedicated training required by LongStream. 

\noindent\textbf{Camera pose estimation.} 
To evaluate the effectiveness of our method in camera pose estimation, we conduct camera pose estimation experiments on ScanNet and TUM dynamics.
\Cref{tab:pose_length} shows camera pose estimation performances across varying sequence lengths. 
Our method consistently achieves the lowest ATE on both datasets, and outperforms all baselines across every metric on ScanNet at 500 and 1,000 frames.
At a sequence length of 1,000 frames, our method reduces the $\text{ATE}$ from 0.39 to 0.18 on ScanNet, achieving approximately a 54\% improvement over TTT3R.

\noindent\textbf{Video depth estimation.} 
We conduct video depth estimation experiments on the Bonn and KITTI datasets across sequence lengths ranging from 50 to 500 frames.
As shown in \Cref{tab:depth_length}, our proposed method significantly outperforms both CUT3R and TTT3R in the long sequence evaluation.
On Bonn, our method reduces $\text{Abs Rel}$ from 0.100 to 0.077 and improves $\delta < 1.25$ from 0.922 to 0.955 over TTT3R at a sequence length of 500.
On KITTI, our method achieves the lowest $\text{Abs Rel}$ across medium and long lengths and improves $\delta < 1.25$ from 0.866 to 0.881 over TTT3R at length 500.
These results demonstrate that our approach also generalizes to indoor and outdoor scenes.

\noindent\textbf{3D reconstruction.}
\begin{figure*}[t]
    \centering
    \includegraphics[width=\textwidth]{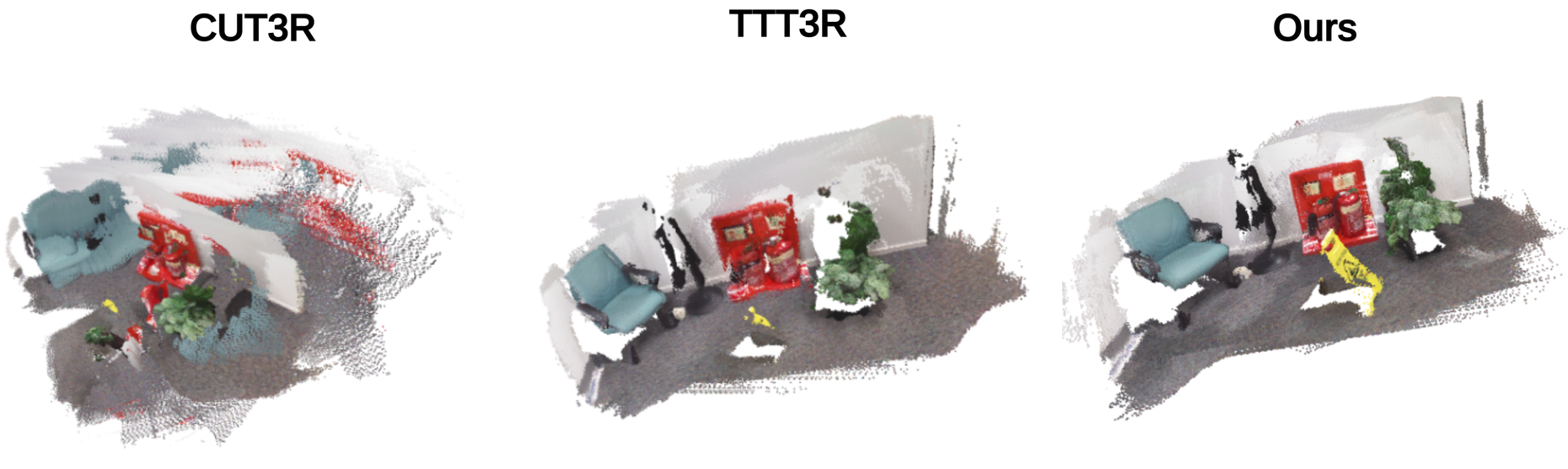}
    \caption[3D Reconstruction Quality on 7-Scenes]{\textbf{3D Reconstruction Quality on 7-Scenes.}}
    \label{fig:qualitative 3d recon 7scenes}
\end{figure*}
\input{tables/3_3d_recon_tables}
We visualize the 3D reconstruction quality on 7-Scenes, and we further report three quantitative metrics on NRGBD and 7-Scenes. 
\Cref{fig:qualitative 3d recon 7scenes} shows qualitative results on the 7-Scenes dataset. Our method demonstrates superior 3D reconstruction quality on both sequences. CUT3R exhibits significant spatial distortion across the scene, while TTT3R achieves better reconstruction overall but still struggles to recover fine-grained details. 
Our method, by contrast, faithfully reconstructs all objects in the scene, as evidenced by the well-preserved shapes of the yellow ladder and the green object in the bottom row.
In terms of quantitative results, our method consistently outperforms the baselines as the sequence length grows, as reported in \Cref{tab:recon_length}. On 7-Scenes, the performance of our method is well preserved even as the number of frames increases, and on NRGBD it maintains a noticeably higher level of accuracy than prior approaches. The gap is even more pronounced on Comp, where our method remains nearly flat while CUT3R and TTT3R deteriorate substantially. While NC is occasionally slightly lower than that of TTT3R, the differences remain marginal across all settings.

\section{Discussion}
\label{sec:analysis}

\subsection{Analysis of State and Model Confidence}
\label{sec:analysis state token}
To analyze how the world-coordinate confidence and state tokens of the model evolve over long sequences, we observe the confidence and the rank of the state tokens throughout extended inputs.
For CUT3R, the confidence gradually decreases as shown in \Cref{fig:confidence}. At the same time, the rank of the state grows rapidly, which can be attributed to the indiscriminate accumulation of redundant information from the input image stream.
TTT3R partially mitigates this rank growth, but the effect remains limited.
Our Information-Aware State Update (\textit{Ours w/o\ reset}), however, substantially slows the rank growth and maintains it at a moderate level.
This result indicates that our method effectively filters out redundant information and selectively incorporates only salient information, thereby preserving the well-established 3D geometric information and leading to enhanced overall performance.
Moreover, our full method (Ours) mitigates the confidence collapse issue and reduces the saturated rank via Dynamic State Reset.
As a result, our method restores plasticity in the long image stream and overcomes the information saturation issue inherent in state tokens.
\input{tables/4_analysis_figure}
\subsection{Robustness of Proposed Method}
\label{sec:robustness}
To examine whether the proposed informativeness term acts as a reliable indicator of input quality, we observe the effective rank ($r_{\text{eff}}$) under the simulated corruption scenarios.
Specifically, we replace the middle 50 frames of each sequence with ImageNet-C corrupted versions \cite{imagenetc}, simulating Motion Blur and Gaussian Noise.
As shown in \Cref{fig:image token rank}, the rank of image tokens significantly decreases at the onset of corruption, demonstrating that the proposed metric effectively identifies less meaningful frames. 
By suppressing the update magnitude during these intervals, the framework keeps the state tokens uncontaminated and preserves the accumulated 3D geometry information.

We further provide quantitative results in \Cref{tab:robustness}. Across both datasets, our method achieves the best performance on nearly all metrics under both Motion Blur and Gaussian Noise. 
In most cases, incorporating $w_\text{info}$ leads to clear performance gains, and even where it does not, Abs Rel remains superior. 
These results confirm that the informativeness-aware update mechanism yields consistent performance under corrupted image streams.

\begin{table}[t]
\centering
\caption[Robustness evaluation under image-level corruptions on Bonn and KITTI]{\textbf{Robustness evaluation under image-level corruptions on Bonn and KITTI.} 
}
\label{tab:robustness}
\resizebox{\linewidth}{!}{
\begin{tabular}{lcc|cc}
\toprule
\multirow{2}{*}{Method} 
& \multicolumn{2}{c|}{Bonn (Length 500)} 
& \multicolumn{2}{c}{KITTI (Length 500)} \\
\cmidrule(lr){2-3} \cmidrule(lr){4-5}
& Abs Rel($\downarrow$) & $\delta<{1.25}(\uparrow)$
& Abs Rel($\downarrow$) & $\delta<{1.25}(\uparrow)$ \\
\midrule
\multicolumn{5}{c}{Motion Blur (level 5)} \\
\midrule
CUT3R               & 0.097 & 91.8 & 0.159 & 77.8 \\
TTT3R               & 0.091 & \underline{94.5} & 0.141 & 83.2 \\
\midrule
Info3R w/o $w_{\text{info}}$ & \underline{0.086} & 93.8 & \underline{0.133} & \textbf{88.3} \\
Info3R (Ours)               & \textbf{0.081} & \textbf{94.9} & \textbf{0.132} & \underline{83.5} \\
\midrule
\multicolumn{5}{c}{Gaussian Noise (level 5)} \\
\midrule
CUT3R               & 0.109 & 88.8 & 0.168 & 76.7 \\
TTT3R               & 0.103 & \textbf{93.0} & 0.149 & 82.6 \\
\midrule
Info3R w/o $w_{\text{info}}$ & \underline{0.101} & 88.7 & \underline{0.138} & \underline{83.7} \\
Info3R (Ours)             & \textbf{0.099} & \underline{90.2} & \textbf{0.133} & \textbf{84.5} \\
\bottomrule
\end{tabular}
}
\end{table}





\subsection{Analysis of Dynamic State Reset}
To analyze when the proposed Dynamic State Reset is triggered, we visualize the reset time steps on Seq.~01 and Seq.~07 of the KITTI Odometry dataset. Seq.~01 consists primarily of straight segments, with a turning segment appearing in the latter part of the sequence. The visualization shows that resets are concentrated in this turning segment. The same tendency is observed in Seq.~07, which contains multiple turns: four regions exhibit notably frequent resets, all of which precisely coincide with turning points along the trajectory. These results indicate that our method triggers resets at moments when the validity of the accumulated scene information diminishes—that is, when viewpoint changes render the information stored in the existing state inconsistent with the new observations. This behavior effectively mitigates the accumulation of irrelevant information that arises when processing long image streams.

\begin{figure}[t]
    \centering
    \includegraphics[width=\linewidth]{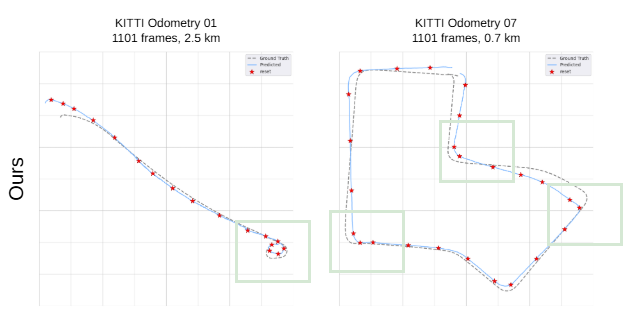}
    \caption[Camera trajectories and reset timings]{\textbf{Camera trajectories and reset timings.}}
    \label{fig:appendix reset timing}
\end{figure}

\subsection{Efficiency Comparison}
\begin{table}[t]
\centering
\caption[Efficiency comparison on KITTI Odometry sequence ``00'']{\textbf{Efficiency comparison on KITTI Odometry sequence ``00''.}}
\label{tab:efficiency}
\small
\setlength{\tabcolsep}{3pt}
\begin{tabular}{lccc}
\toprule
Method & Mem. (GB, $\downarrow$) & FPS ($\uparrow$) & ATE ($\downarrow$) \\
\midrule
CUT3R      & \textbf{3.23}  & 15.27          & 188.49 \\
TTT3R      & 4.70           & 15.17          & 180.00 \\
LongStream & 10.77          & \textbf{20.47} & 77.05  \\
\midrule
Info3R (Ours)       & 3.24           & 13.27          & \textbf{30.03} \\
\bottomrule
\end{tabular}
\end{table}
We measure the runtime speed and peak GPU memory consumption on KITTI Odometry Seq.~00 using a single NVIDIA RTX 4090 GPU.
\Cref{tab:efficiency} reports the memory consumption, runtime, and accuracy of each method on KITTI Odometry Seq.~00. In terms of memory, our method maintains identical memory consumption to the base model CUT3R, introducing only a marginal overhead. 
On the other hand, TTT3R incurs additional storage cost during decoding due to its cross-attention-based alignment confidence, while LongStream consumes more than $3\times$ the memory of our method due to its causal-attention design.
In terms of inference speed, our method runs slightly slower than baselines because of the additional adaptive update and reset operations. 
Nevertheless, it achieves the lowest ATE among all compared methods. Overall, our method offers a favorable trade-off between efficiency and accuracy.

\subsection{Ablation Study}
\begin{table}[t]
\centering
\caption[Ablation study on KITTI Odometry]{\textbf{Ablation study on KITTI Odometry.}}
\label{tab:ablation_trigger}
\small
\setlength{\tabcolsep}{3pt}
\begin{tabular}{lccc}
\toprule
Variants & ATE ($\downarrow$) & $\text{RPE}_{\text{rot}}$ ($\downarrow$) & $\text{RPE}_{\text{trans}}$ ($\downarrow$) \\
\midrule
w/o $w_{\text{redundant}}$ & 45.03 & \textbf{0.239} & \textbf{0.213} \\
w/o $w_{\text{info}}$      & 32.47 & 0.283 & 0.430 \\
\midrule
Reset w/o $\alpha_t$                  & 36.68 & 0.347 & 0.490 \\
Reset w/o $\bar{C}_t^{\text{world}}$  & 36.18 & 0.393 & 0.932 \\
\midrule
Info3R (Ours) & \textbf{30.48} & 0.333 & 0.663 \\
\bottomrule
\end{tabular}
\end{table}
We conduct ablation studies on the KITTI Odometry dataset. Our method consists of two components, and we ablate each individually in \Cref{tab:ablation_trigger}. For the Information-Aware State Update, the redundancy weight has a larger impact than the informativeness weight. Removing $w_\text{redundant}$ degrades ATE from 30.48 to 45.03, whereas removing $w_\text{info}$ yields an ATE of 32.47. This result highlights the importance of preventing redundant updates to the state tokens.
Notably, while removing $w_\text{redundant}$ slightly improves RPE, 
it substantially degrades ATE, indicating that the redundancy weight primarily 
contributes to global trajectory consistency rather than local frame-to-frame accuracy.
For the Dynamic State Reset, the two trigger signals contribute comparably in terms of ATE: 36.68 without $\alpha_t$ and 36.18 without $\bar{C}_t^{\text{world}}$.
Jointly tracking information accumulation and model stability yields the best performance, confirming that both signals are necessary for our reset mechanism.

\section{CONCLUSIONS}
We propose Information-adaptive test-time training for 3D reconstruction (Info3R), addressing two limitations of TTT3R: the lack of frame-wise importance weighting and of a proper state reset for long image streams. Info3R introduces (1) Information-Aware Update, which adjusts the learning rate by each frame's importance, and (2) Dynamic State Reset, which mitigates state saturation and confidence collapse in long sequences. 
Our training-free method outperforms TTT3R on camera pose estimation, video depth estimation, and 3D reconstruction, and surpasses the additionally-trained LongStream on the long-sequence KITTI Odometry benchmark. 
We further find that declining world-coordinate confidence coincides with information saturation (rising state-token rank), motivating our design.




\section*{APPENDIX}
\subsection{Hyperparameters of Proposed Method}
\label{sec:hyperparameters}
We report the dataset-specific hyperparameters used in our experiments in \Cref{tab:hyperparameters}. Our method has two main hyperparameters: the reset trigger threshold $\gamma$, which controls how aggressively the state token is reset based on the accumulated confidence-weighted update, and the base learning rate $\alpha_{\text{base}}$, which scales the adaptive update of the state token. We additionally fix $\tau = 100$ across all datasets, as the median number of image tokens per frame measured across datasets falls within the range of 75--100. We use the same configuration for all sequences within each dataset.
\begin{table}[h]
\centering
\caption[Hyperparameters of our method on each dataset]{\textbf{Hyperparameters of our method on each dataset.} 
$\gamma$: reset trigger threshold; 
$\alpha_{\text{base}}$: base learning rate. 
$\tau=100$ across all datasets.}
\label{tab:hyperparameters}
\small
\setlength{\tabcolsep}{8pt}
\resizebox{\linewidth}{!}{
\begin{tabular}{ll|cc}
\toprule
Task & Dataset & $\gamma$ & $\alpha_{\text{base}}$ \\
\midrule
Long sequence pose estimation & KITTI Odometry  & 3.2  & 1.7 \\
\midrule
\multirow{2}{*}{Pose estimation} 
        & ScanNet         & 3  & 1.0 \\
        & TUM             & 20 & 1.0 \\
\midrule
\multirow{2}{*}{Depth estimation}
        & Bonn            & 3  & 1.0 \\
        & KITTI           & 0.3  & 0.4 \\
\midrule
\multirow{2}{*}{3D reconstruction}
        & 7-Scenes        & 25 & 0.4 \\
        & NRGBD           & 10 & 0.5 \\

\bottomrule
\end{tabular}
}
\end{table}

\bibliographystyle{IEEEtran}
\bibliography{references}

\end{document}

%% file: tables/0_long_camera_pose.tex
\begin{table*}[t]
\centering
\caption[Quantitative comparison on the KITTI Odometry dataset]{\textbf{Quantitative comparison on the KITTI Odometry dataset.}}
\label{tab:kitti_result}
\footnotesize 
\setlength{\tabcolsep}{3pt}
\resizebox{\linewidth}{!}{
\begin{tabular}{l ccccc ccccc c c}
\toprule
\multirow{3}{*}{\textbf{Methods}} & \multicolumn{11}{c}{\textbf{KITTI Odometry~\cite{kitti}} ($\text{ATE}$ $\downarrow$)} & \multirow{3}{*}{\textbf{Avg.}($\downarrow$)} \\
\cmidrule(lr){2-12}
& \makecell{00 \\ \tiny 4542x, 3.7km} & \makecell{01 \\ \tiny 1101x, 2.5km} & \makecell{02 \\ \tiny 4661x, 5.1km} & \makecell{03 \\ \tiny 801x, 0.6km} & \makecell{04 \\ \tiny 271x, 0.4km} & \makecell{05 \\ \tiny 2761x, 2.2km} & \makecell{06 \\ \tiny 1101x, 1.2km} & \makecell{07 \\ \tiny 1101x, 0.7km} & \makecell{08 \\ \tiny 4071x, 3.2km} & \makecell{09 \\ \tiny 1591x, 1.7km} & \makecell{10 \\ \tiny 1201x, 0.9km} & \\
\midrule

CUT3R & 187.79 & 638.10 & 279.32 & 153.64 & 22.98 & 152.89 & 132.52 & 73.47 & 233.12 & 176.99 & 185.36 & 203.29 \\
TTT3R & 161.60 & 537.34 & 263.74 & 106.43 & 11.91 & 145.46 & 128.47 & 69.33 & 236.65 & 181.62 & 127.32 & 179.08 \\
LongStream & 77.05 & \textbf{49.45} & 169.20 & \textbf{3.66} & \textbf{2.18} & 75.22 & 12.83 & 15.66 & 61.60 & 77.38 & 19.34 & 51.24 \\
\midrule
Info3R (Ours) & \textbf{30.03}& 97.20 & \textbf{86.57} & 9.11 & 6.26 & \textbf{23.60} & \textbf{11.21} & \textbf{7.64} & \textbf{29.99} & \textbf{22.89} & \textbf{10.76} & \textbf{30.48} \\
\bottomrule
\end{tabular}
}
\end{table*}

%% file: tables/0_visualize_long_camera_pose.tex
\begin{figure*}[t]
    \centering
    \includegraphics[width=\linewidth]{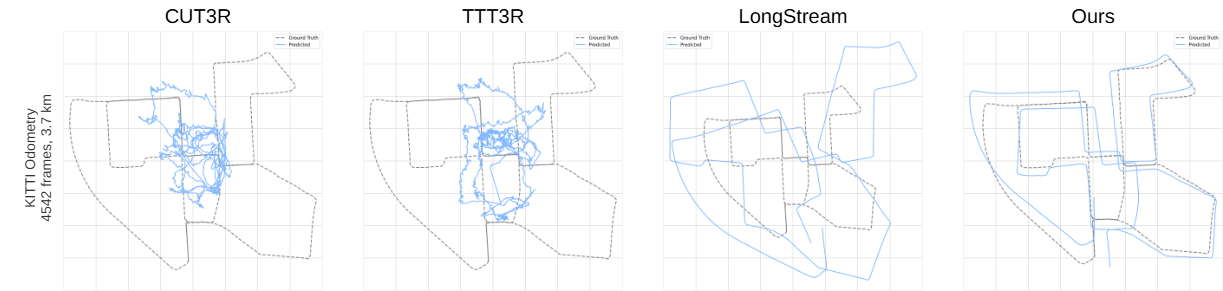}
    \caption[Camera trajectories on Seq.~00 in KITTI Odometry]{\textbf{Camera trajectories on Seq.~00 in KITTI Odometry.}}
    \label{fig:kitti traj error}
\end{figure*}

%% file: tables/1_camera_pose_tables.tex
\begin{table*}[t]
\centering
\setlength{\tabcolsep}{2pt} 
\caption[Camera Pose estimation across different sequence lengths]{\textbf{Camera Pose estimation across different sequence lengths.}
For each setting, $\text{RPE}_\text{T}$ ($\text{RPE}_\text{trans}$) and $\text{RPE}_\text{R}$ ($\text{RPE}_\text{rot}$) are reported in the second and third columns of each block, respectively.
}
\label{tab:pose_length}
\resizebox{\textwidth}{!}{
\begin{tabular}{lccc|ccc|ccc|ccc|ccc|ccc}
\toprule
\multirow{3}{*}{Method} 
& \multicolumn{9}{c|}{TUM} 
& \multicolumn{9}{c}{ScanNet} \\
\cmidrule(lr){2-10} \cmidrule(lr){11-19}
& \multicolumn{3}{c|}{Short (50)} 
& \multicolumn{3}{c|}{Medium (500)} 
& \multicolumn{3}{c|}{Long (1000)} 
& \multicolumn{3}{c|}{Short (50)} 
& \multicolumn{3}{c|}{Medium (500)} 
& \multicolumn{3}{c}{Long (1000)} \\
\cmidrule(lr){2-4} \cmidrule(lr){5-7} \cmidrule(lr){8-10} \cmidrule(lr){11-13} \cmidrule(lr){14-16} \cmidrule(lr){17-19}
& ATE($\downarrow$) & RPE$_\text{T}$($\downarrow$) & RPE$_\text{R}$($\downarrow$)
& ATE($\downarrow$) & RPE$_\text{T}$($\downarrow$) & RPE$_\text{R}$($\downarrow$)
& ATE($\downarrow$) & RPE$_\text{T}$($\downarrow$) & RPE$_\text{R}$($\downarrow$)
& ATE($\downarrow$) & RPE$_\text{T}$($\downarrow$) & RPE$_\text{R}$($\downarrow$)
& ATE($\downarrow$) & RPE$_\text{T}$($\downarrow$) & RPE$_\text{R}$($\downarrow$)
& ATE($\downarrow$) & RPE$_\text{T}$($\downarrow$) & RPE$_\text{R}$($\downarrow$) \\
\midrule
CUT3R & 0.026 & 0.009 & 0.327 & 0.151 & \textbf{0.008} & 0.417 & 0.183 & \textbf{0.008} & 0.544
      & 0.045 & 0.018 & 0.478 & 0.674 & 0.034 & 1.161 & 0.825 & 0.035 & 1.203 \\
TTT3R & 0.015 & \textbf{0.008} & 0.310 & 0.073 & 0.011 & \textbf{0.373} & 0.117 & 0.010 & \textbf{0.480}
      & \textbf{0.033} & \textbf{0.017} & \textbf{0.450} & 0.271 & 0.036 & 1.096 & 0.394 & 0.039 & 1.162 \\
\midrule
Info3R (Ours)  & \textbf{0.014} & \textbf{0.008} & \textbf{0.305} & \textbf{0.036} & 0.010 & 0.377 & \textbf{0.061} & 0.011 & 0.508
      & \textbf{0.033} & 0.018 & 0.489 & \textbf{0.141} & \textbf{0.022} & \textbf{0.596} & \textbf{0.176} & \textbf{0.023} & \textbf{0.595} \\
\bottomrule
\end{tabular}
}
\end{table*}

%% file: tables/2_video_depth_tables.tex
\begin{table*}[t]
\centering
\caption[Video Depth estimation across different sequence lengths]{\textbf{Video Depth estimation across different sequence lengths.} 
}
\label{tab:depth_length}
\resizebox{\textwidth}{!}{
\begin{tabular}{lcc|cc|cc|cc|cc|cc}
\toprule
\multirow{3}{*}{Method} 
& \multicolumn{6}{c|}{Bonn} 
& \multicolumn{6}{c}{KITTI} \\
\cmidrule(lr){2-7} \cmidrule(lr){8-13}
& \multicolumn{2}{c|}{Short (50)} 
& \multicolumn{2}{c|}{Medium (250)} 
& \multicolumn{2}{c|}{Long (500)} 
& \multicolumn{2}{c|}{Short (50)} 
& \multicolumn{2}{c|}{Medium (250)} 
& \multicolumn{2}{c}{Long (500)} \\
\cmidrule(lr){2-3} \cmidrule(lr){4-5} \cmidrule(lr){6-7} \cmidrule(lr){8-9} \cmidrule(lr){10-11} \cmidrule(lr){12-13}
& Abs Rel$(\downarrow)$ & $\delta_{1.25}(\uparrow)$
& Abs Rel$(\downarrow)$ & $\delta_{1.25}(\uparrow)$
& Abs Rel$(\downarrow)$ & $\delta_{1.25}(\uparrow)$
& Abs Rel$(\downarrow)$ & $\delta_{1.25}(\uparrow)$
& Abs Rel$(\downarrow)$ & $\delta_{1.25}(\uparrow)$
& Abs Rel$(\downarrow)$ & $\delta_{1.25}(\uparrow)$ \\
\midrule
CUT3R & 0.106 & 0.897 & 0.104 & 0.882 & 0.099 & 0.907
      & 0.112 & 0.876 & 0.129 & 0.851 & 0.151 & 0.806 \\
TTT3R & 0.087 & 0.946 & 0.103 & 0.900 & 0.100 & 0.922
      & \textbf{0.105} & \textbf{0.893} & 0.112 & \textbf{0.896} & 0.132 & 0.866 \\
\midrule
Info3R (Ours)  & \textbf{0.071} & \textbf{0.966} & \textbf{0.073} & \textbf{0.960} & \textbf{0.077} & \textbf{0.955}
      & 0.115 & 0.869 & \textbf{0.107} & 0.891 & \textbf{0.116} & \textbf{0.881} \\
\bottomrule
\end{tabular}
}
\end{table*}

%% file: tables/3_3d_recon_tables.tex
\begin{table*}[t]
\centering
\setlength{\tabcolsep}{2pt} 
\caption[3D reconstruction across different sequence lengths]{\textbf{3D reconstruction across different sequence lengths.} 
}
\label{tab:recon_length}
\resizebox{\textwidth}{!}{
\begin{tabular}{lccc|ccc|ccc|ccc|ccc|ccc}
\toprule
\multirow{3}{*}{Method} 
& \multicolumn{9}{c|}{NRGBD} 
& \multicolumn{9}{c}{7-Scenes} \\
\cmidrule(lr){2-10} \cmidrule(lr){11-19}
& \multicolumn{3}{c|}{Short (50)} 
& \multicolumn{3}{c|}{Medium (200)} 
& \multicolumn{3}{c|}{Long (400)} 
& \multicolumn{3}{c|}{Short (50)} 
& \multicolumn{3}{c|}{Medium (200)} 
& \multicolumn{3}{c}{Long (400)} \\
\cmidrule(lr){2-4} \cmidrule(lr){5-7} \cmidrule(lr){8-10} \cmidrule(lr){11-13} \cmidrule(lr){14-16} \cmidrule(lr){17-19}
& Acc$(\downarrow)$ & Comp$(\downarrow)$ & NC$(\uparrow)$
& Acc$(\downarrow)$ & Comp$(\downarrow)$ & NC$(\uparrow)$
& Acc$(\downarrow)$ & Comp$(\downarrow)$ & NC$(\uparrow)$
& Acc$(\downarrow)$ & Comp$(\downarrow)$ & NC$(\uparrow)$
& Acc$(\downarrow)$ & Comp$(\downarrow)$ & NC$(\uparrow)$
& Acc$(\downarrow)$ & Comp$(\downarrow)$ & NC$(\uparrow)$ \\
\midrule
CUT3R & 0.041 & 0.016 & 0.673 & 0.133 & 0.032 & 0.600 & 0.314 & 0.109 & 0.554
      & 0.021 & 0.019 & 0.607 & 0.087 & 0.049 & 0.564 & 0.163 & 0.102 & 0.533 \\
TTT3R & 0.034 & 0.015 & \textbf{0.680} & 0.063 & 0.013 & \textbf{0.625} & 0.141 & 0.071 & 0.591
      & \textbf{0.018} & 0.019 & \textbf{0.610} & 0.027 & 0.023 & \textbf{0.581} & 0.049 & 0.026 & 0.558 \\
\midrule
Info3R (Ours) & \textbf{0.031} & \textbf{0.014} & 0.665 & \textbf{0.043} & \textbf{0.009} & 0.619 & \textbf{0.072} & \textbf{0.018} & \textbf{0.614}
      & \textbf{0.018} & \textbf{0.018} & 0.609 & \textbf{0.019} & \textbf{0.021} & 0.579 & \textbf{0.020} & \textbf{0.020} & \textbf{0.559} \\

\bottomrule
\end{tabular}
}
\end{table*}

%% file: tables/4_analysis_figure.tex
\begin{figure*}[t]
    \centering
    \begin{subfigure}[t]{0.32\linewidth}
        \centering
        \includegraphics[width=\linewidth]{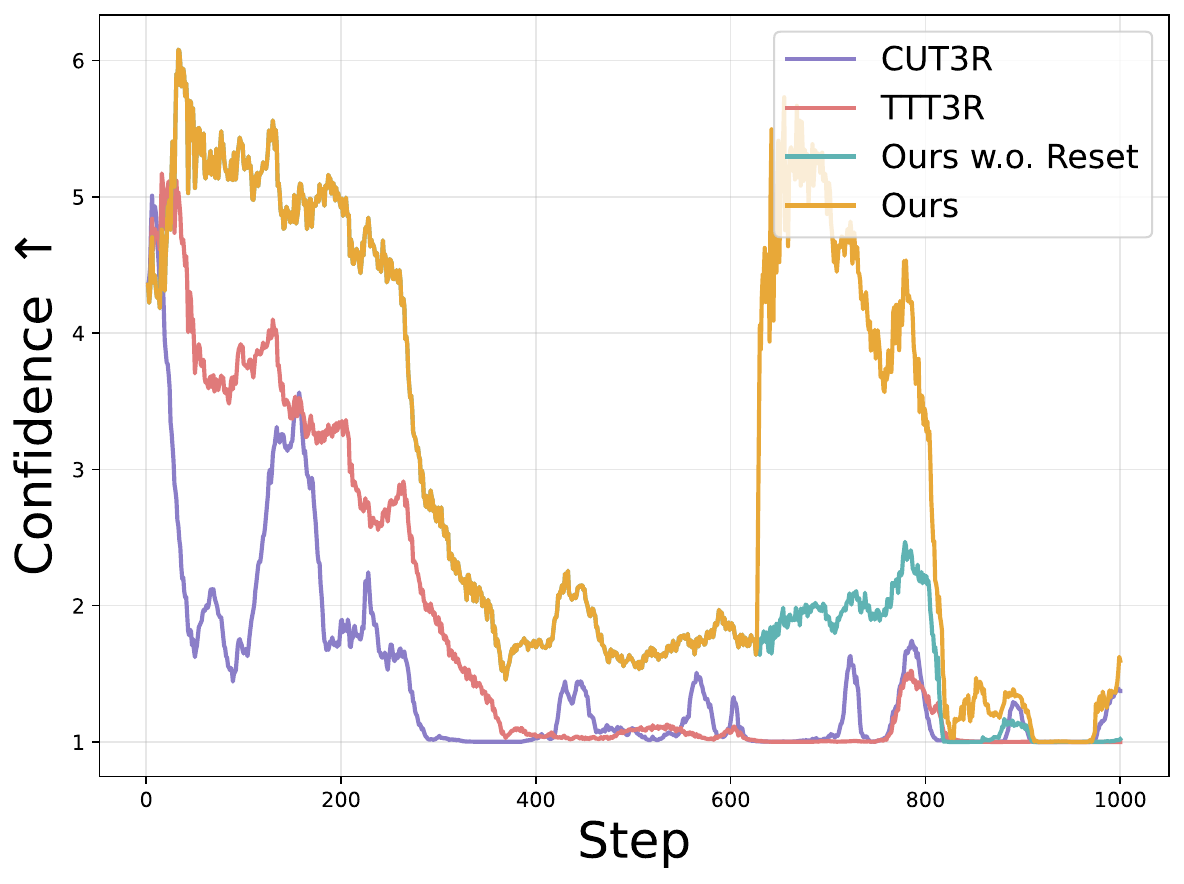}
        \caption{Comparison of model confidence.}
        \label{fig:confidence}
    \end{subfigure}
    \hfill
    \begin{subfigure}[t]{0.32\linewidth}
        \centering
        \includegraphics[width=\linewidth]{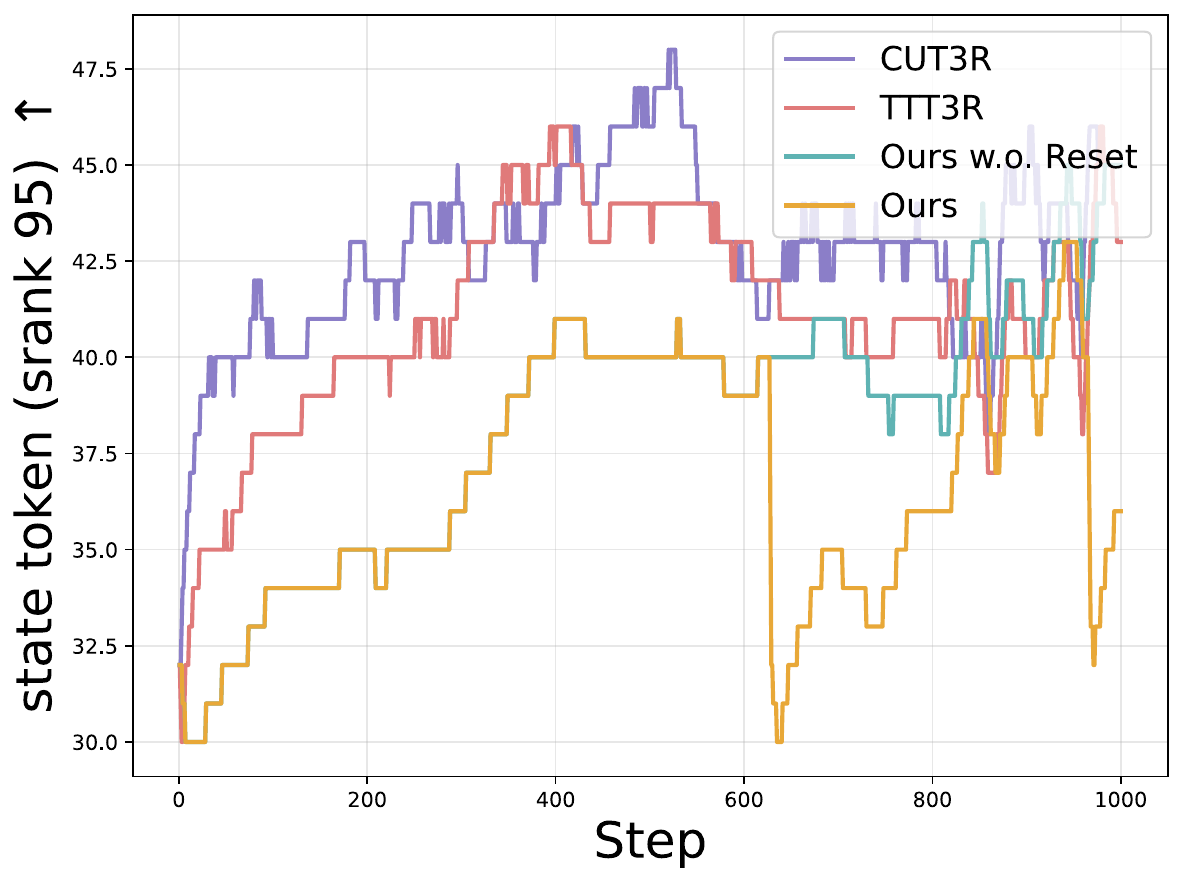}
        \caption{Rank of the state.}
        \label{fig:state token rank}
    \end{subfigure}
    \hfill
    \begin{subfigure}[t]{0.32\textwidth}
        \centering
        \includegraphics[width=\linewidth]{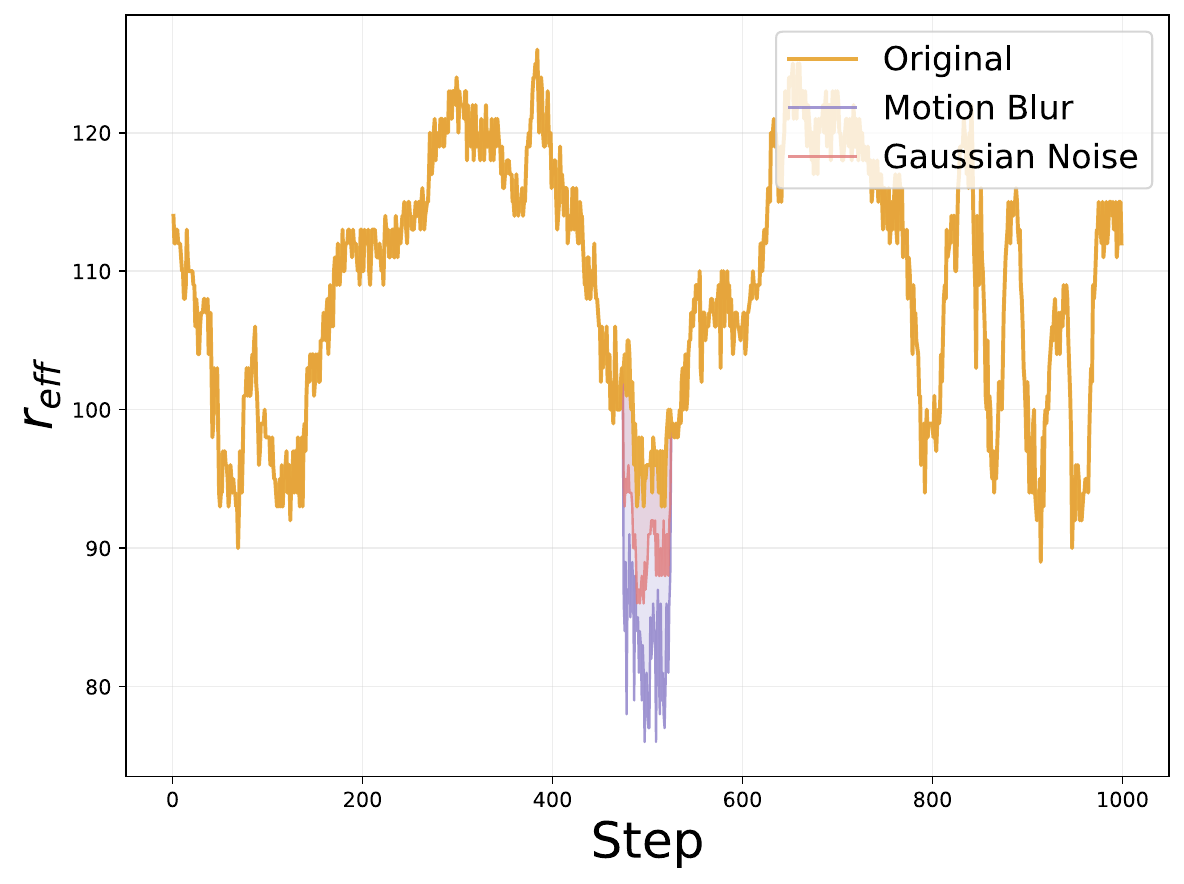}
        \caption{The effective rank $r_\text{eff}$ under motion blur and Gaussian noise.}
        \label{fig:image token rank}
    \end{subfigure}
    \caption[Analyses on the TUM dynamics dataset]{\textbf{Analyses on the TUM dynamics dataset.}}
    \label{fig:analyis}
\end{figure*}